\documentclass{article}

\PassOptionsToPackage{sort, numbers}{natbib}

\usepackage[final]{neurips_data_2023}

\usepackage[utf8]{inputenc} 
\usepackage[T1]{fontenc}    
\usepackage{hyperref}       
\usepackage{url}            
\usepackage{amsfonts}       
\usepackage{nicefrac}       
\usepackage{microtype}      
\usepackage{xcolor}         
\usepackage{graphicx} 
\usepackage{tabularx}
\usepackage{makecell}
\usepackage{longtable}

\usepackage{arydshln}
\usepackage{siunitx}
\usepackage{amsmath}
\usepackage{amssymb}
\usepackage{booktabs}       

\usepackage[inkscapelatex=false]{svg}

\usepackage{subfigure}

\usepackage{siunitx} 
\usepackage{soul}
\usepackage{multicol}
\usepackage{multirow}

\newcolumntype{x}[1]{>{\centering\arraybackslash\hspace{0pt}}m{#1}}

\newenvironment{remark}[1][Remark]{\begin{trivlist}
\item[\hskip \labelsep {\bfseries #1}]}{\end{trivlist}}

\newcommand{\nop}[1]{}

\newcommand{\mind}{\textsc{Mind2Web}}
\newcommand{\model}{\textsc{MindAct}}

\newcommand\eg[0]{\textit{e.g.}}
\newcommand\ie[0]{\textit{i.e.}}

\title{\mind: Towards a Generalist Agent for the Web}

\author{%
  Xiang Deng\thanks{Corresponding authors: \{deng.595, sun.397, su.809\}@osu.edu} \quad Yu Gu \quad Boyuan Zheng \quad Shijie Chen\\
  \textbf{Samuel Stevens \quad Boshi Wang \quad Huan Sun$^*$ \quad Yu Su$^*$}\\
  The Ohio State University\\
  \url{https://osu-nlp-group.github.io/Mind2Web} \\
}

\begin{document}

\maketitle

\begin{abstract}

We introduce \mind, the first dataset for developing and evaluating generalist agents for the web that can follow language instructions to complete complex tasks on any website.
Existing datasets for web agents either use simulated websites or only cover a limited set of websites and tasks, thus not suitable for generalist web agents.
With over \num{2000} open-ended tasks collected from \num{137} websites spanning \num{31} domains and crowdsourced action sequences for the tasks, \mind\ provides three necessary ingredients for building generalist web agents: 1) diverse domains, websites, and tasks, 2) use of real-world websites instead of simulated and simplified ones, and 3) a broad spectrum of user interaction patterns. 
Based on \mind, we conduct an initial exploration of using large language models (LLMs) for building generalist web agents.
While the raw HTML of real-world websites are often too large to be fed to LLMs, we show that first filtering it with a small LM significantly improves the effectiveness and efficiency of LLMs. 
Our solution demonstrates a decent level of performance, even on websites or entire domains the model has never seen before, but there is still a substantial room to improve towards truly generalizable agents.
We open-source our dataset, model implementation, and trained models ({{\small \url{https://osu-nlp-group.github.io/Mind2Web}}}) to facilitate further research on building a generalist agent for the web. 
\end{abstract}

\section{Introduction}

\begin{figure}
    \centering
    \includegraphics[width=\textwidth]{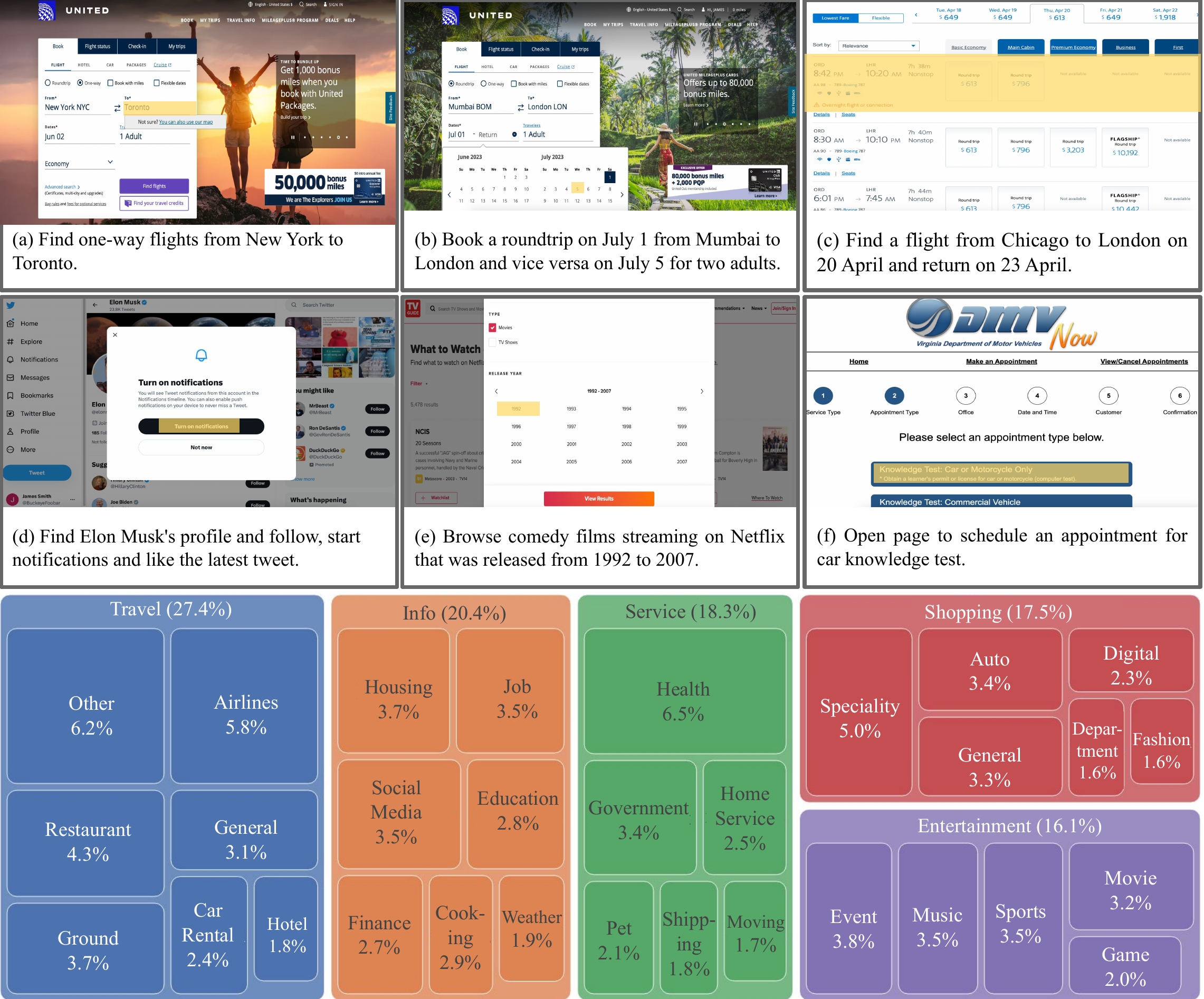}
    \caption{Sample tasks and all domains featured in \mind. The array of diversity allows for testing an agent's generalizability across tasks on the same website (a vs.\ b), similar tasks on different websites (a vs.\ c), and even to entirely disparate tasks, websites, and domains (d$-$f).}
    \label{fig:data_overall}
    \vspace{-1em}
\end{figure}

The web now hosts billions of websites~\cite{howmany} that cover virtually every aspect of the digital world. 
In this work, we seek to answer the question: \textit{How can we build a generalist agent for the web that, given any website, can follow language instructions and carry out the corresponding tasks?}
Some exemplar tasks for such an agent are shown in Figure~\ref{fig:data_overall}.
A generalist agent could make the web more accessible, which is becoming increasingly difficult as modern websites provide increasingly more functionalities that also increase their complexity and learning curve.  
On the other hand, such an agent may also turn the entire web into an unprecedentedly powerful and versatile tool~\cite{augmented,tool-survey} that can enhance large language models (LLMs). 
For example, it may be used as a plugin for ChatGPT~\cite{chatgpt-plugins} to directly acquire information and carry out actions on HTML websites, instead of only retrieving web content through a retriever tool~\cite{chen-etal-2017-reading,karpukhin-etal-2020-dense} or relying on pre-defined APIs for each web service~\cite{nl2api, taskcompletionflows}.

A generalist agent for the web shall meet the following desiderata:
First, \textbf{it shall work on any website on the Internet}. 
Since it is infeasible to collect sufficient training data that covers all websites, this requires the agent to be inherently generalizable to websites or even domains it has never seen before. 
Second, \textbf{it shall work on real-world websites, which can be dynamic, complex, and noisy}.
Most modern websites are dynamic, generating and rendering different content in response to user actions. 
This necessitates the agent to model each website as a partially-observable environment instead of assuming full knowledge \textit{a priori}. 
The agent should also not make strong simplifying assumptions about the environments but must embrace the full complexity and sometimes noise, \eg, due to sub-optimal website designs. 
Finally, \textbf{it shall support diverse and sophisticated interactions with websites.}
Tasks from users can be highly diverse and take a large number of steps to complete (\eg, the task in Figure~\ref{fig:data_overall}(b) would take \num{14} actions).
An agent that only supports simple tasks may provide limited value to users.

Building an agent for the web is not entirely new.
There have been numerous prior efforts in varied forms. 
However, none of them meets all the requirements for a generalist agent listed above. 
Existing work falls short in one to all of the following aspects: 1) Only operating in a limited and pre-specified set of websites~\cite{yaoWebShopScalableRealWorld2022, sunMETAGUIMultimodalConversational2022, Liu2018ReinforcementLO, Li2020MappingNL, Burns2022ADF}, 2) making strong simplifying assumptions about the websites~\cite{Liu2018ReinforcementLO, yaoWebShopScalableRealWorld2022}, and 3) only supporting specific types of tasks~\cite{yaoWebShopScalableRealWorld2022, Li2020MappingNL, Liu2018ReinforcementLO} and/or requiring tedious step-by-step instructions from users~\cite{Liu2018ReinforcementLO, Xu2021GroundingOI, Li2020MappingNL, Burns2022ADF}. 
Meanwhile, LLMs have been shown to excel at grounded language understanding in complex environments with good generalizability and sample efficiency~\cite{pangu, SayCan, llm-planner, structgpt}.
It is therefore promising to explore LLMs as a candidate solution towards generalist agents for the web.
However, there lacks a good dataset that can support the development and evaluation of generalist web agents, which is the focus of this work.

In light of this, we present \mind, a new dataset with natural language tasks and manually annotated action sequences for developing and evaluating generalist agents for the web. 
It offers the following unique features:

\textbf{1. Diverse coverage of domains, websites, and tasks.} \mind\ boasts a collection of over \num{2000} tasks curated from \num{137} websites that span \num{31} different domains. This extensive range of tasks and domains not only provides a vast landscape for exploration and learning but also opens up a new level of versatility and complexity, fostering a more comprehensive evaluation of generalist web agents.

\textbf{2. Use of real-world websites.} \mind\ replaces the oversimplified simulation environments commonly found in other datasets with an authentic, vibrant, and unpredictable realm of real-world websites. We provide full traces of user interactions, webpage snapshots, and network traffic, making it a rich source of raw, unfiltered, and dynamic data. By doing so, \mind\ equips models with the capacity to interact and cope with the complexities and uncertainties of real-world environments, thereby encouraging the development of more robust and adaptive models.

\textbf{3. A broad spectrum of user interaction patterns.} \mind\ enables users to engage in sophisticated ways with websites, as opposed to basic operations such as searching, following links, and reading content commonly found in existing work. Users can \textit{click, select, and type in any elements} on the website, which significantly expands the space of possible tasks. This captures all the common actions users do on websites in real life and promotes the development of agents capable of handling complex tasks.

With the diverse domains and websites, we create challenging out-of-distribution evaluation settings where agents are tested on their generalizability to websites or even entire domains never seen during training.
This presents a representative evaluation of generalist agents working on unseen websites.

\mind\ enables us to conduct an initial exploration in using LLMs for building generalist web agents.
The HTML document of real-world webpages may contain thousands of elements, which makes it infeasible or too costly to be fed into an LLM's context. To address this issue, we propose \model, a two-stage model that involves first using a fine-tuned small LM to filter the web elements and then using an LLM to select from the filtered elements in a multi-choice question answering fashion and predict the corresponding action with the selected element.
We show that \model\ significantly outperforms modeling strategies commonly adopted by prior work and achieves a decent level of generalization.
It can also work well with both open-source LLMs like Flan-T5~\cite{flant5} through fine-tuning and closed-source LLMs like GPT-3.5-turbo and GPT-4 through in-context learning.
However, there is still substantial room for further improvement towards generalist agents for the web.
Promising future directions include integrating multi-modal information, reinforcement learning with feedback from real websites, and specialized LMs for web understanding and action taking.

\section{\mind\ Dataset}

Unlike existing datasets predominantly constructed within simulated environments~\cite{shiWorldBitsOpenDomain2017, yaoWebShopScalableRealWorld2022}, our objective is to bridge the gap between simulation and reality so that agents trained on our dataset can work on real-world websites out of the box.
To achieve this, our approach for data collection adheres to the following principles. 
Firstly, instead of recreating websites in simulation, which often leads to oversimplified environments, we engage directly with real-world websites and capture snapshots of these environments.
Secondly, we collate a diverse set of websites from varied domains and crowdsource realistic tasks that cover a wide range of functionalities provided by these websites.
Finally, acknowledging the challenge of perfectly replicating the complexity of real-world environments, we strive to capture a comprehensive snapshot of each website and the full interaction trace, to the extent that all the tasks can be seamlessly replayed offline. This supports rich modeling and evaluation approaches, ensuring a robust and practical dataset for research.

\subsection{Task Definition}
The primary objective of \mind\ is for the agent to complete a specific task on the target website through a series of actions. Each instance in our dataset contains three components:

\noindent\textbf{Task description,} which outlines the high-level goal of the task. We intentionally avoid low-level, step-by-step instructions, aiming to foster the development of agents that can comprehend and carry out tasks in a more autonomous fashion, rather than merely following prescriptive directives.

\noindent\textbf{Action sequence}, which is the sequence of actions required to accomplish the task on the website. Each action in the sequence comprises a \texttt{(Target Element, Operation)} pair. The \texttt{Target Element} is an interactable element on the current webpage, and the \texttt{Operation} refers to the action to be executed on that element. We support three common operations: \texttt{Click} (also including \texttt{Hover} and \texttt{Press Enter}), \texttt{Type}, and \texttt{Select Option}. For \texttt{Type} and \texttt{Select Option}, they also require an additional value as an argument.
Actions in a sequence often span multiple webpages of a website.

\noindent\textbf{Webpage snapshots}, which constitute the environment within which the task is performed. We provide the snapshots in a variety of formats to accommodate different modeling approaches:
self-contained MHTML file that includes the raw HTML code of the webpage,
DOM snapshot containing the DOM tree along with the layout and style information of the screenshot of the rendered webpage,
HAR file that includes all the network traffic for replaying the interaction if needed,
and trace file that comprises the complete interaction trace during the task annotation process.

The agent receives the task description in the beginning.
At each step, it also receives the current webpage and the history of previous actions. The objective is to accurately predict the subsequent action, which encompasses the target element for interaction and the operation.

\begin{figure}
    \centering
    \includegraphics[width=\textwidth]{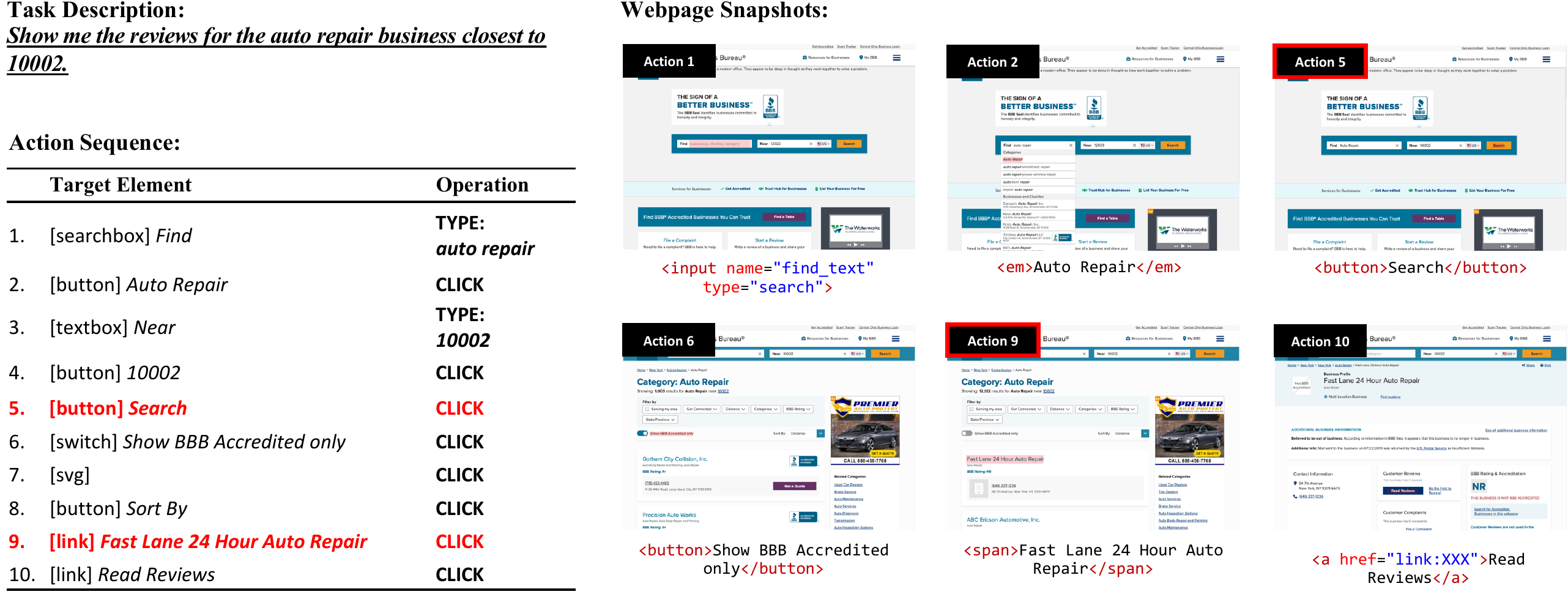}
    \caption{A sample data instance of our dataset with the three components. Actions marked in \textcolor{red}{\textbf{red}} will result in a transition to a new webpage.}
    \label{fig:data_example}
    \vspace{-.5em}
\end{figure}

\subsection{Data Collection}
Our data collection process consists of four stages: website selection, task proposal, task demonstration, and task verification.
Website selection and task verification are done by the authors.
For task proposal and demonstration, we develop a sophisticated annotation tool using Playwright~\footnote{\url{https://playwright.dev/}} and hire annotators through Amazon Mechanical Turk.
Refer to Supplementary for annotation tool details.

\noindent\textbf{Website Selection.}
We start with \num{5} top-level domains: Travel, Shopping, Service, Entertainment, and Information, which are subsequently broken down into \num{31} (secondary) domains.
We select websites within each domain based on their popularity in the US, as ranked by similarweb.com.
We manually select \num{3}-\num{5} representative websites per domain, resulting in a collection of \num{137} websites in total.

\noindent\textbf{Task Proposal.}
We present the annotators with a target website, a concise description of the website, and a few sample tasks associated with it.
The annotators are then asked to propose \textit{open-ended and realistic tasks} based on three criteria: the tasks should be of diverse types, require multiple rounds of interaction, and describe the high-level goal instead of step-by-step instructions.
To further stimulate creativity and boost diversity, we use ChatGPT to generate seed tasks by prompting it to test different functionalities of a website.
We generate \num{50} seed tasks per website, of which \num{10} are randomly sampled and presented to the annotator each time.
These seed tasks are mainly for inspiration---annotators are explicitly instructed not to directly use them and we reject task proposals that are highly similar to the seed tasks.
All the proposed tasks are further screened by the authors to ensure quality and diversity before entering the demonstration phase.

\noindent\textbf{Task Demonstration.}
We develop a Playwright-based tool for demonstration (Figure~\ref{fig:data_example}).
Workers will use the tool to demonstrate how to perform the tasks they have proposed within a web browser.
To ensure accuracy, each interaction round is split into two parts: \textit{element selection} and \textit{operation selection}.
At each step, the worker first selects an element on the webpage by clicking within the browser. They are then asked to confirm the selection and choose the operation to execute on the selected element.
Once the task is completed, the worker is given another opportunity to review and modify the task description.

\noindent\textbf{Task Verification.}
Lastly, all task demonstrations are verified by the authors to ensure the following:
First, all actions are accurately reflected in the task description. The authors will modify the task description if needed to align it with the annotated actions;
Second, the recorded actions are correct and clean, with extraneous steps discarded.
Finally, the starting and ending points of tasks are consistent, such as excluding actions for closing popup windoes, or ending the annotation at the search result page if the task was to find a certain item without clicking on specific items.
After verification, we discarded \num{61} out of the total \num{2411} tasks. Among the \num{2350} retained tasks, the task description was refined in \num{390} instances to better correspond with the demonstrated actions, while some extraneous steps were discarded in \num{187} instances. Overall, the data collection pipeline has been proven effective and produces high-quality data.

\begin{table}
    \centering
    \caption{Statistics of \mind\ compared with existing datasets.}
    \resizebox{\textwidth}{!}{
    \begin{tabular}{lccx{2.55cm}cx{2.5cm}x{3cm}c}
    \toprule
    &\textbf{\# Dom.} &
    \textbf{\# Env.} &
    \textbf{Env. Type} &
    \textbf{\makecell{Avg. \#\\Elements}} &
    \textbf{\# Tasks} &  \textbf{Task Info.} &
    \textbf{\makecell{Avg. \#\\ Actions}} \\
    \midrule
       MiniWoB++~\cite{Liu2018ReinforcementLO}&$-$&\num{100}&Simplified \quad mobile websites&\num{28}&\num{100}&Low-level&\num{3.6}\\
       WebShop~\cite{yaoWebShopScalableRealWorld2022}&\num{1}&\makecell{\num{1}}&Simplified shopping websites&\num{38}&\num{12000} products&High-level&\num{11.3}\\
       RUSS~\cite{Xu2021GroundingOI}&$-$&\num{22}&Real-world websites&\num{801}&\num{80}&\makecell{High \& low}&\num{5.4}\\
       PixelHelp~\cite{Li2020MappingNL}&\num{4}&\num{4}&Mobile apps&$-$&\num{187}&\makecell{High \& low}&-\\
       META-GUI~\cite{sunMETAGUIMultimodalConversational2022}&\num{6}&\num{11}&Mobile apps&\num{79}&\makecell{\num{1125} dialogues}&\makecell{High-level}&\num{4.3}\\
       MoTIF~\cite{Burns2022ADF}&\num{15}&\num{125}&Mobile apps&\num{188}&\num{756}&\makecell{High \& Low}&\num{4.4}\\
       \midrule
        \mind & \num{5} / \num{31}&\num{137}&Real-world websites&\num{1135}&\num{2350}&\makecell{High-level}&\num{7.3}\\
        \bottomrule
    \end{tabular}
    }
    \vspace{-.5em}
    \label{tab:cmp}
\end{table}
\subsection{Comparison with Existing Work and Research Challenges}

\mind\ presents a unique ensemble of research challenges for the development of generalist agents for the web in real-world settings.
As shown in Table~\ref{tab:cmp}, \mind\ distinguishes itself from existing literature in several ways.
Firstly, \mind\ spans across \num{137} websites from \num{31} domains, allowing comprehensive testing of an agent's ability in generalizing across varied environments.
Secondly, we utilize real-world websites without manual simplification. Consequently, the included environments exhibit complexity far surpassing that encountered in previous studies, yet better reflecting the intricacy of the modern web. With an average of over \num{1000} elements per page embedded within complex DOM structures, how to effectively process such long and highly structured documents presents a significant challenge for modeling.
Lastly, we direct the annotators to propose \textit{open-ended tasks} that explore different functionalities of the website to mimic genuine web usage.
Meanwhile, contrary to prior studies~\cite{Liu2018ReinforcementLO,Xu2021GroundingOI,Li2020MappingNL,Burns2022ADF} that provide step-by-step directives and primarily focus on testing the agent's ability to translate low-level instructions into actions, \eg, ``\textit{Type New York in the location field, click the search button and choose the tomorrow tab},'' we opted for the setting where only high-level goals are available, \eg, ``\textit{What is the weather for New York tomorrow}?'' This poses a much greater yet realistic planning and grounding challenge for the agent.

\section{Method: \model}
Employing the data from \mind, we introduce an exploratory framework, \model, for our task, leveraging the power of LLMs.
Raw HTML documents, which could consist of thousands of elements, are either infeasible or cost-prohibitive to be directly fed into LLMs.
We propose a two-stage process that synergizes the strength of small and large LMs, as shown in Figure~\ref{fig:overview}.
In the first stage, a fine-tuned small LM is used to rank the elements present on a webpage, yielding a small pool of promising candidates.
In the second stage, these candidate elements are consolidated to form a representative snippet of the webpage, which is then processed by an LLM to predict the final action, including predicting both the element for interaction and the corresponding operation.

\subsection{Candidate Generation with Small LMs}
\begin{figure}
    \begin{minipage}{.52\linewidth}
       \centering
        \includegraphics[width=0.9\textwidth]{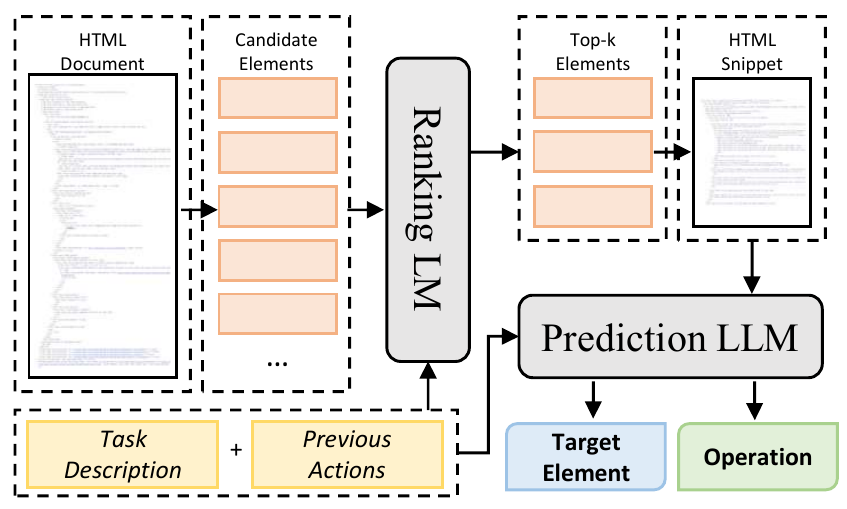}
        \caption{The overall pipeline for \model\ with a small ranking LM for candidate generation, and a large prediction LM for action prediction.}
        \label{fig:overview}
    \end{minipage}%
    \hfill
    \begin{minipage}{.44\linewidth}
        \centering
        \includegraphics[width=0.85\textwidth]{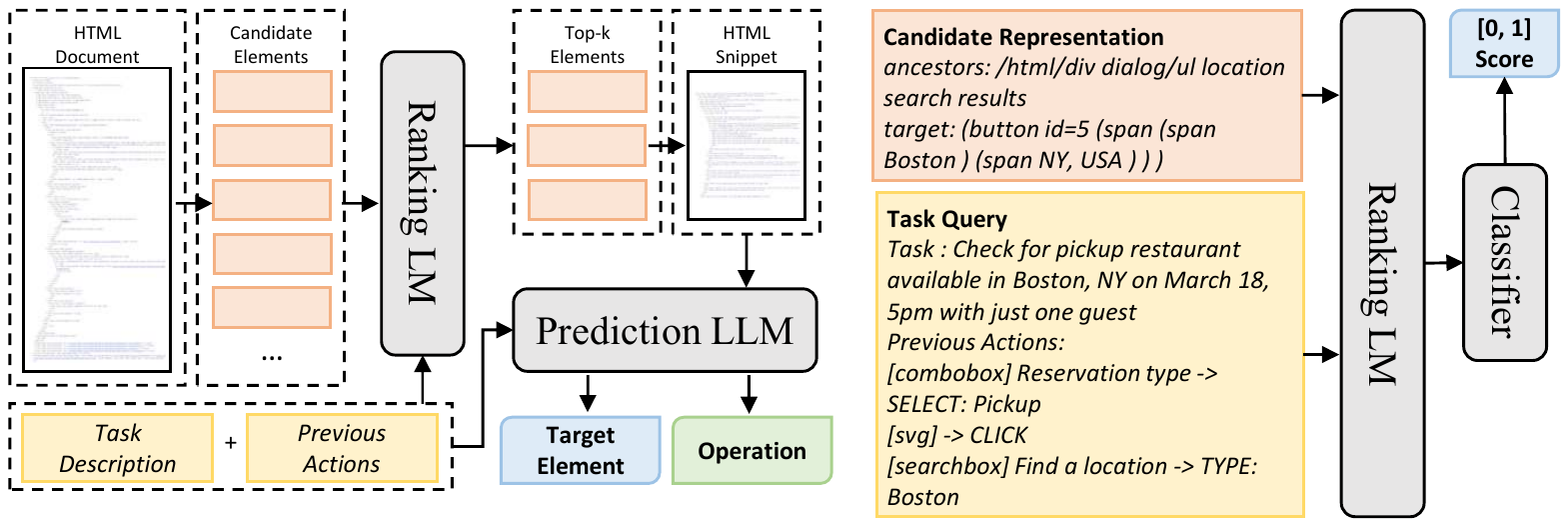}
        \caption{Illustration of the candidate generation module and the templates for constructing task query and candidate representation.}
        \label{fig:candidate_generation}
    \end{minipage} 
    \vspace{-1em}
\end{figure}
Given the task description, the snapshot of the webpage at step $t$, and the actions performed in the preceding $t-1$ steps, we treat candidate generation as a ranking task. The task is to select the top-$k$ candidate DOM elements from the webpage that best align with both the task description and the current step.
We formulate the task query by concatenating the task description with previous actions.
The textual representation of each candidate DOM element is derived from a combination of the element's tag, its textual content, and salient attribute values, as well as the textual representation of its parent and child elements.
As shown in Figure~\ref{fig:candidate_generation}, we pair each DOM element with the task query and feed it to an encoder-only LM through the cross-encoder architecture~\cite{reimers-2019-sentence-bert}, yielding a matching score.
At training time, we randomly sample negative elements from the webpage, and use the target element as the positive example. The matching score is passed through a sigmoid activation function and optimized with a binary cross entropy loss. At inference time, we score all elements in the webpage and pick the top-$k$ elements with the largest logits as input to the second stage.

\subsection{Action Prediction with LLMs}
\begin{figure}
    \centering
    \includegraphics[width=0.9\textwidth]{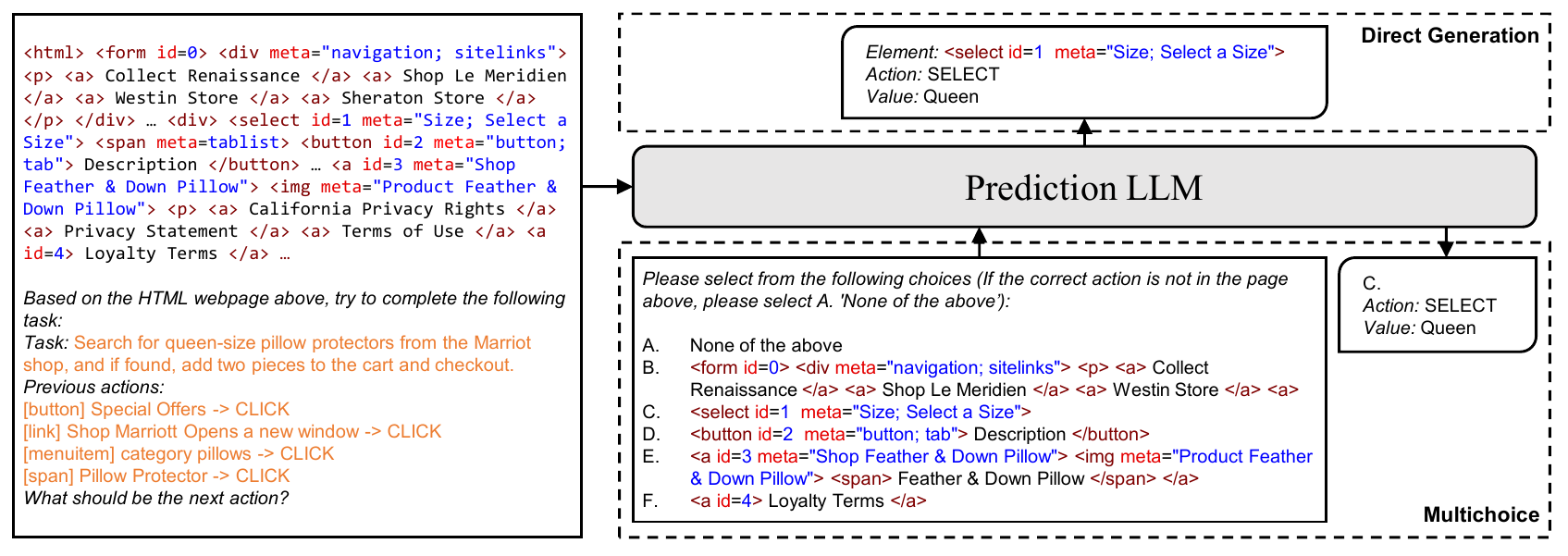}
    \caption{Illustration of action prediction with LLMs.}
    \label{fig:template}
    \vspace{-1.5em}
\end{figure}

After obtaining the top-$k$ candidates, we utilize the candidate set to prune the webpage snapshot and construct snippets that only include the selected candidates and their neighbours as inputs to an LLM.
Recent studies~\cite{flant5,pangu} have suggested that training LMs for discrimination rather than generation is more generalizable and sample-efficient for other grounding tasks.
Inspired by that, we convert the task of element selection into a \textit{multi-choice question answering} (QA) problem. 
Instead of generating the complete target element, the LM is trained to instead select from a list of options.
For comparison, we also include a baseline that directly generates the target element based on the provided webpage snippet.
In both cases, we directly let the LLM generate the operation, along with the additional value needed for some operations.
An example is shown in Figure~\ref{fig:template}.
We incorporate up to \num{5} candidate elements within each input, together with a None option, and partition the candidate set into several groups.
During training, we construct the target sequence using ground-truth actions and fine-tune the model using a left-to-right language modeling objective.
During inference, we divide the top-$k$ candidates into multiple clusters of five options. If more than one option is selected after a round, we form new groups with the selected ones. This process repeats until a single element is selected, or all options are rejected by the model, i.e., the model chooses the None option for all groups.

\section{Experiments}

\subsection{Experimental Setup}
\label{sec:data_split}
The diversity of \mind\ provides a unique opportunity to evaluate an agent's generalizability at different levels.
We seek to understand how well an agent can generalize across domains, websites, and tasks:
\textbf{Test\textsubscript{Cross-Domain}},
for which we hold out two top-level domains, \textit{Information} and \textit{Service}, with \num{912} tasks from \num{73} websites. Here, the model is expected to generalize to an entirely new domain without having seen any websites or tasks associated with that domain during training.
\textbf{Test\textsubscript{Cross-Website}},
with \num{10} websites from each remaining top-level domain, containing \num{177} tasks. In this setting, the model is never exposed to the test websites during training. However, it has been trained on websites from the same domain and possibly with similar tasks. 
This setup allows us to assess an agent's ability to adapt to entirely new websites, yet within familiar domains and task contexts.
\textbf{Test\textsubscript{Cross-Task}},
where we randomly split \num{20}\% of the remaining data, regardless of domains and websites, resulting in \num{252} tasks from \num{69} websites. In this setting, the model has been exposed to webpages from the same website during training and has likely encountered similar tasks.
The rest of data is used for training, which contains \num{1009} tasks from \num{73} websites.

\subsection{Data Preprocessing and Evaluation}
\label{sec:evaluation}
We apply simple heuristics to clean the raw HTML documents, keeping only elements that are visible and carry substantial semantic meaning, as determined by their attributes, textual content, and neighboring elements. This effectively reduces the average number of elements from \num{1135} to \num{580}, while still maintaining an overall recall of \num{94.7}\% for the target element in the training data.

For evaluation, we first calculate \textbf{Element Accuracy} that compares the selected element with all acceptable elements, and \textbf{Operation F1} that calculates token-level F1 score for the predicted operation. This is the same as accuracy for \texttt{Click}, but considers the correctness of the input value for \texttt{Type} and \texttt{Select Option}. Each step of the task is evaluated independently with the ground truth action history provided. We then define \textbf{Step Success Rate} and \textbf{Success Rate} (for the whole task). A step is regarded as successful only if both the selected element and the predicted operation are correct. A task is regarded successful only if all steps have succeeded. It is therefore a stringent metric. For step-wise metrics, we report macro average across tasks.

\begin{table}[t]
\centering
\caption{Main results. The classification baseline uses DeBERTa\textsubscript{B} and the generation baseline uses Flan-T5\textsubscript{B}. For step-wise metrics, we report macro average across tasks. $^*$ For GPT-4 we use \num{50} tasks for each setting with top-\num{10} candidates due to limited budget. See Appendix~\ref{app:result_50} for results on the \num{50} tasks subsets for all methods.}
\resizebox{\textwidth}{!}{
\begin{tabular}{lcccccccccccc}
\toprule
& \multicolumn{4}{c}{\textbf{Cross-Task}} & \multicolumn{4}{c}{\textbf{Cross-Website}} & \multicolumn{4}{c}{\textbf{Cross-Domain}}\\
     \cmidrule(lr){2-5}\cmidrule(lr){6-9}\cmidrule(lr){10-13}
     &\textbf{Ele. Acc}&\textbf{Op. F1}&\textbf{Step SR}&\textbf{SR}&\textbf{Ele. Acc}&\textbf{Op. F1}&\textbf{Step SR}&\textbf{SR}&\textbf{Ele. Acc}&\textbf{Op. F1}&\textbf{Step SR}&\textbf{SR}\\
     \midrule
     Classification& \num{26.8}&$-$&$-$&$-$&\num{21.6}&$-$&$-$&$-$&\num{24.5}&$-$&$-$&$-$ \\
     \midrule
     \multirow{1}{*}{Generation}
     & \num{20.2}&\num{52.0}&\num{17.5}&\num{0.0}&\num{13.9}&\num{44.7}&\num{11.0}&\num{0.0}&\num{14.2}&\num{44.7}&\num{11.9}&\num{0.4}\\
     \midrule
     \model\\
     \hspace{1em}w/ Flan-T5\textsubscript{B}& \num{43.6}&\num{76.8}&\num{41.0}&\num{4.0}&\num{32.1}&\textbf{\num{67.6}}&\num{29.5}&\num{1.7}&\num{33.9}&\textbf{\num{67.3}}&\num{31.6}&\num{1.6}\\
     \hspace{1em}w/ Flan-T5\textsubscript{L}& \num{53.4}&\textbf{\num{75.7}}&\num{50.3}&\textbf{\num{7.1}}&\num{39.2}&\num{67.1}&\num{35.3}&\num{1.1}&\num{39.7}&\num{67.2}&\num{37.3}&\num{2.7}\\
     \hspace{1em}w/ Flan-T5\textsubscript{XL}& \textbf{\num{55.1}}&\textbf{\num{75.7}}&\textbf{\num{52.0}}&\num{5.2}&\textbf{\num{42.0}}&\num{65.2}&\textbf{\num{38.9}}&\textbf{\num{5.1}}&\textbf{\num{42.1}}&\num{66.5}&\textbf{\num{39.6}}&\textbf{\num{2.9}}\\
     \addlinespace[0.1em]\hdashline\addlinespace[0.1em]
     \hspace{1em}w/ GPT-3.5&\num{20.3}&\num{56.6}&\num{17.4}&\num{0.8}&\num{19.3}&\num{48.8}&\num{16.2}&\num{0.6}&\num{21.6}&\num{52.8}&\num{18.6}&\num{1.0} \\
     \hspace{1em}w/ GPT-4$^*$&\num{41.6}&\num{60.6}&\num{36.2}&\num{2.0}&\num{35.8}&\num{51.1}&\num{30.1}&\num{2.0}&\num{37.1}&\num{46.5}&\num{26.4}&\num{2.0} \\
\bottomrule
\end{tabular}}
\label{tab:main}
\end{table}
\subsection{Results}
\noindent\textbf{Candidate Generation.}
We fine-tune DeBERTa~\cite{deberta} as the small LM for candidate generation. As candidate generation requires high efficiency, we use the base version DeBERTa\textsubscript{B} with \num{86}M parameters. Overall, it achieves \num{88.9}\% / \num{85.3}\% / \num{85.7}\% Recall@\num{50} on  Test\textsubscript{Cross-Task}, Test\textsubscript{Cross-Website} and Test\textsubscript{Cross-Domain}, respectively. We use its top-\num{50} ranking results as the candidate pool for all subsequent experiments. 

\noindent\textbf{Action Prediction.}
We mainly compare against two baselines in Table~\ref{tab:main}.
The first directly uses the candidate generation model (DeBERTa) for element selection, which is similar to existing work~\cite{sunMETAGUIMultimodalConversational2022, gur2023understanding} that combines an encoder with classification heads. However, such a design cannot benefit from many recent LMs that use an encoder-decoder or decoder-only architecture.
It cannot predict actions and the element selection performance is also not competitive, as shown in Table~\ref{tab:main}.
We use Flan-T5~\cite{flant5} as the backbone for the generation model.
The autoregressive generation formulation (Figure~\ref{fig:template} top) does not perform well, and even underperforms the classification baseline on element selection despite the larger model size (\num{220}M for Flan-T5\textsubscript{B}).
We observe a substantial gain with \model\ using the multi-choice QA formulation. The best model achieves \num{52.0}\% step success rate under Cross-Task setting, and \num{38.9}\% / \num{39.6}\% when generalizing to unseen websites and domains.
However, the overall task success rate remains low for all models, as the agent often commits at least one error step in most cases.

\begin{figure}
        \centering
        \includegraphics[width=\linewidth]{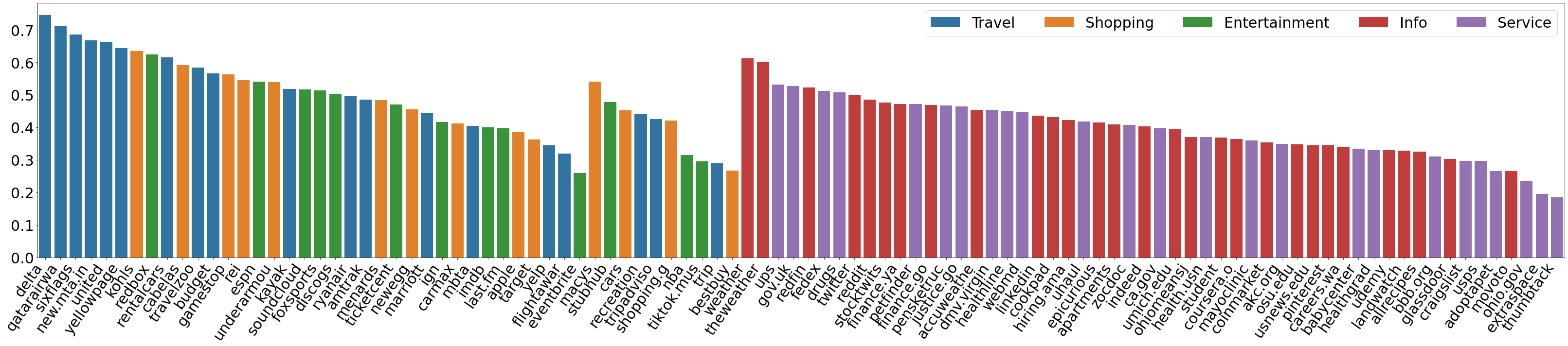}
        \caption{Step success rate per website grouped by the three splits: Test\textsubscript{Cross-Task}, Test\textsubscript{Cross-Website} and Test\textsubscript{Cross-Domain} from left to right. Here we only show websites with more than three test tasks.}
        \label{fig:website_results}
        \vspace{-.5em}
\end{figure}

\noindent\textbf{Three Levels of Generalization.} All models perform best on the Cross-Task setting, with over \num{10}\% absolute gap (step SR) on average compared with Cross-Website and Cross-Domain settings, indicating that generalizing to unseen environments is still a major challenge.
On the contrary, we note that the performance of Cross-Website and Cross-Domain settings are notably similar, which is also reinforced in Figure~\ref{fig:website_results}, where there is no clear distinction in performance across these settings.
This suggests that the challenges primarily stem from the diversity in website designs and interaction logic rather than domain specifics.
Tasks across domains tend to share common operations, and pretrained LMs may already have the capability to decompose complex tasks at a high level based on commonsense knowledge. Yet, grounding such knowledge into actionable steps in specific and varying environments remains a considerable challenge.

\noindent\textbf{In-context Learning with LLMs.}
We also experiment with two popular LLMs, GPT-3.5-turbo and GPT-4~\cite{openai2023gpt4}, through in-context learning. We use the same multiple-choice formulation as \model, and include three demonstration examples for in-context learning.  
We can see that both models are comparable to the two baselines with only three in-context examples.
Note that this is not a fair comparison with the Flan-T5 models, which are fine-tuned on the full training data. We also include the zero-shot results with Flan-T5\textsubscript{XL} in Appendix~\ref{app:flan-t5_zero}, but the model fails to perform the task without fine-tuning.
Meanwhile, GPT-3.5 only has around \num{20}\% element selection accuracy, despite the superior performance people have observed on other datasets.
Further analysis reveals that one possible problem is the model's propensity to select the None option, asserting that the task cannot be finished on the current webpage.
This is somewhat accurate since tasks typically necessitate navigation through multiple webpages and performing a series of actions before reaching the final result. This aspect indeed represents the primary difficulty of our task.
On the other hand, we observe highly promising outcomes with GPT-4. The performance is on par with the tuned Flan-T5 models under Cross-Website and Cross-Domain settings for element selection, indicating a great potential for developing generalist agents using LLMs.
Nevertheless, GPT-4's high operational cost remains a concern. Developing smaller models specialized for the web is an interesting future avenue.
\section{Related Work}
\begin{remark} [Autonomous Agents for Web and Mobile Applications.]
Considerable initiatives have been invested in automating web navigation, driven by a vision of facilitating effortless human-web interaction.
Yet, previous research has been limited by the types of tasks and websites it can handle, either confined to simplified simulation environments~\cite{shiWorldBitsOpenDomain2017, Liu2018ReinforcementLO, yaoWebShopScalableRealWorld2022}, or limited to a narrow set of domains and websites~\cite{yaoWebShopScalableRealWorld2022, Xu2021GroundingOI}.
Recent studies~\cite{sunMETAGUIMultimodalConversational2022, Burns2022ADF, Li2020MappingNL} have utilized similar techniques for mobile applications, however, these are often simpler and offer fewer functions compared with full-fledged websites.
In contrast, \mind\ aims to adapt to a realistic web environment, characterized by its high diversity.
Also related is the research on web automation systems~\cite{leshedCoScripterAutomatingSharing2008, puppeteer}.
These technologies often demand programming skills, which can make them less accessible to general users. We aim to equip the web automation system with a natural language interface, thereby reducing the entry barrier significantly.
\end{remark}

\begin{remark} [Large Language Models.]
In recent years, there has been a surge in the development and application of large language models (LLMs). These models, often encompassing billions of parameters, are pre-trained on massive corpora of text data~\cite{Zhou2023ACS, LLMSurvey, Bommasani2021FoundationModels}, enabling them to capture intricate linguistic patterns, nuances, and relationships, resulting in unprecedented performance on a wide array of NLP tasks.
One of the most noteworthy attributes of LLMs is their few-shot learning capability. Unlike traditional machine learning models that necessitate extensive labeled data for task-specific fine-tuning, LLMs can often perform tasks with minimal task-specific examples. Furthermore, LLMs such as GPT-3~\cite{brownLanguageModelsAre2020} and PaLM~\cite{chowdhery2022palm} have also demonstrated the ability to do in-context learning, where they can adapt to novel tasks by simply providing context within the input prompt, eliminating the need for explicit retraining.
In this work, we explore the use of LLMs to build generalist agent on top of \mind\ by either tuning medium-sized LMs with only around \num{1000} examples, or prompting an LLM such as GPT-4, and have observed promising results.

\end{remark}

\begin{remark} [Grounded Language Understanding.]
Our work also aligns with the field of grounded language understanding, which aims to map natural language utterances onto executable plans in a target environment~\cite{pangu}. 
Many studies have centered around environments underpinned by a well-structured schema or ontology, including relational databases~\cite{spider, wangRATSQLRelationAwareSchema2020} and knowledge bases~\cite{yihSemanticParsingStaged2015,gu2021beyond}, which may not adequately reflect the more heterogeneous conditions in real-world situations.
Our work instead grounds natural language in the noisy and schemaless web environment.
Our setting is also connected to embodied AI, where an agent, guided by language instructions, carries out tasks in a physical environment~\cite{alfred, SayCan, llm-planner}.
Nonetheless, existing research primarily focuses on a specific setting (e.g., household environments), limiting their diversity.
\mind\ provides a unique testbed for studying a broad range of grounding challenges in real-world environments.
\end{remark}

\begin{remark} [Tool Learning.]
Recent developments have underscored LLMs' potential in using a myriad of tools (\ie, taking actions) to augment their capacity~\cite{augmented, tool-survey}, including search engine, translator, calculator, etc.
Example works include Toolformer~\cite{toolformer}, ReAct~\cite{react}, and ToolkenGPT~\cite{toolkengpt}.
The creation of recent benchmarks on tool learning~\cite{api-bank, gorilla} further highlights the growing interest in evaluating LLMs' proficiency in tool usage.
However, existing research primarily concentrates on short-term tool invocation, neglecting long-term planning. 
\mind\ can bridge this lacuna by necessitating LLMs to take actions within realistic web-browsing environments that demand prolonged decision-making sequences. 
Furthermore, \mind\ may stimulate the development of more advanced tools based on LLMs that interface the web with natural language\nop{, thereby obfuscating the low-level intricacies such as handling a DOM object}. 
These advanced tools could be subsequently employed by another LLM for more challenging problem-solving tasks~\cite{openagi, hugginggpt}.
\end{remark}
\section{Limitations and Potential Societal Impact}
\label{app:limitation}
\mind\ is designed to facilitate the development and evaluation of generalist agents for the web. Such agents hold great potential for making the web more accessible and easy to use, especially for individuals who are less familiar with information technology or have disabilities and may struggle to navigate through complex web apps and get overwhelmed by the options available.
However, there are still potential concerns and limitations regarding the current data collection, system design and safety for deployment in real world.

\noindent\textbf{Diversity and Representation in Data Collection.}
Although we strive to choose representative websites covering diverse domains, the present selection predominantly comprises English-language websites primarily used in the U.S.
Meanwhile, all our annotators are sourced through the Amazon MTurk platform, which might be biased towards a group that is more proficient in web use. 
Therefore, the tasks and websites embodied in our dataset may represent only a subset of all potential tasks that can be performed on the web.
Bearing this limitation in mind, the design of \mind\ and our data collection protocol allow for easy expansion to encompass more tasks and websites.
The inclusion of additional websites, potentially from different countries and languages, and tasks from more diverse demographics, such as individuals from different age groups, those traditionally facing web accessibility challenges, and professionals from specific domains like software development, research, law, and more, present exciting directions for future development.

\noindent\textbf{Use of Multimodal Information.}
Our current approach, \model, models the web environment using only textual context from webpage snapshots. Nevertheless, crucial information can also be gleaned from the visual representation of a rendered webpage.
While not currently utilized, we have included complete webpage snapshots in \mind, enabling rendering of the webpage for visual interpretation. The use of this multimodal information will be a viable prospect for improving model performance.

\noindent\textbf{Modeling of Interaction Dynamics.}
In \model, we encode each webpage independently at every step, with only the previous actions provided as historical context.
However, the changes of the web environment could also provide significant cues for task completion, such as the appearance of a dropdown menu following a button click.
Exploring effective ways to model such dynamic environment transformations during interaction could be an essential aspect for developing robust web agents.

\noindent\textbf{Human-Agent Interaction.}
In the current design of \mind, the user provides a single description of the task goal up front, and the agent carries out the task from start to finish.
In real-world settings, the user may wish to adjust or add task requirements in the middle, or the agent might seek user confirmation for more accurate task understanding.
Extending Mind2Web to an interactive or conversational setting, thereby allowing diverse forms of human-agent interactions, could be an interesting future direction.

\noindent\textbf{Evaluation with Offline/Online Environments.}
Following recent works~\cite{sunMETAGUIMultimodalConversational2022, Burns2022ADF}, we evaluate the system with cached offline environments, which allows us to test using snapshots of complex real-world websites.
However, a downside to this is that the task will fail immediately if an action was not cached during data collection, potentially leading to false negatives due to the existence of multiple paths for completing the same task.
As described in Appendix~\ref{app:eval_normalization}, we normalize the actions to address equivalent elements within the same page. 
In addition, we include complete network traffic in the dataset, presenting possibilities for future research to enable some degree of replay and exploration within the cached environment.
Given that \mind\ faithfully replicates real-world webpages, systems trained on the dataset should be readily transferable to live websites. 
Conducting end-to-end live evaluation on real websites with human assistance is a very promising direction that is worth exploration.

\noindent\textbf{Safety in Deployment.}
While the development of general-purpose web agents holds great potential to enhance efficiency, optimize user experiences, and promote web accessibility universally, the accompanying safety considerations for real-world deployment cannot be ignored.
These include how to effectively manage sensitive actions like financial transactions, enhancing transparency and interpretability, and keeping users in control during task execution.
Additionally, there is the risk of these agents possessing the capability to breach existing security measures such as CAPTCHA and being exploited for malicious activities, such as disseminating false information.
Therefore, it is also important for cybersecurity research to consider these potential uses and develop preemptive protective measures.

\section{Conclusion}

In this work, we introduced \mind, the first dataset for developing and evaluating generalist agents for the web. 
We also proposed \model, an agent that leverages the power of (large) language models for effectively tackling this task. 
Our work opens up a wide range of promising future directions, including integrating multi-modal information, reinforcement learning with feedback from real websites, and specialized LMs for web understanding and action taking.
We hope that \mind\ will serve as a valuable platform for the research community to advance towards generalist agents for the web.

\section*{Acknowledgements}
The authors would thank colleagues from the OSU NLP group for constructive feedback and all contributors from the Amazon Mechanical Turk platform who participated in our study and assisted in data collection. 
This research was sponsored in part by NSF OAC 2112606, NSF CAREER \#1942980, ARL W911NF2220144 and Ohio Supercomputer Center~\cite{OhioSupercomputerCenter1987}. The views and conclusions contained herein are those of the authors and should not be interpreted as representing the official policies,
either expressed or implied, of the U.S. government. The U.S. Government is authorized to reproduce and distribute reprints for Government purposes notwithstanding any copyright notice herein.

\bibliographystyle{plainnat}
\bibliography{ref.bib}
\newpage
\appendix

\section{Overview}

Our supplementary includes the following sections:
\begin{itemize}
    \item \textbf{Section~\ref{app:data}: Data Collection Details.} Details for crowsourcing and implementation details for the three data collection phases: task proposal, task demonstration, and task verification.
    \item \textbf{Section~\ref{app:exp}: Experiment Details.} Details for evaluation and model implementation.
    \item \textbf{Section~\ref{app:add_result}: Additional Results.} Results for additional auxiliary experiments.
\end{itemize}

Following NeurIPS Dataset and Benchmark track guidelines, we have shared the following artifacts:

\begin{table}[h]
    \centering
    \begin{tabular}{lll}
    \toprule
    \textbf{Artifcat}     &  \textbf{Link} & \textbf{License}\\
    \midrule
      Homepage   & \url{https://osu-nlp-group.github.io/Mind2Web/} &-\\\addlinespace
      Code Repository & \url{https://github.com/OSU-NLP-Group/Mind2Web} & MIT License\\\addlinespace
      Training Data & \url{https://huggingface.co/datasets/osunlp/Mind2Web} & CC BY 4.0\\\addlinespace
      Test Data & \url{https://shorturl.at/iGI45} (password ``mind2web'') & CC BY 4.0\\
         \bottomrule
    \end{tabular}
    \label{tab:artifact_link}
\end{table}

The authors and the OSU NLP group are committed to ensuring its regular upkeep and updates.

\newpage
\section{Data Collection Details}
\label{app:data}
\subsection{Crowdsourcing Details}
\begin{table}[]
    \centering
    \small
    \caption{Prompt for ChatGPT to generate seed tasks to inspire the annotators during task proposal.}
    \begin{tabular}{lp{10cm}}
    \toprule
      \textbf{Prompt}   &  
      {We are testing the website\newline\newline
      https://aa.com | American Airlines - Airline tickets and low fares at aa.com\newline
      Book low fares to destinations around the world and find the latest deals on airline tickets, hotels, car rentals and vacations at aa.com. As an AAdantage member you earn miles on every trip and everyday spend.\newline\newline
      We want to test practical daily tasks that a user would do on the website.\newline
      Come up with a list of 5 example tasks and try to cover different cases.\newline
      Requirements:\newline\newline
    \ - Each example should be a single sentence and not just click one of the elements.\newline
    \ - Don't give step-by-step instructions or directly mention the element to interact.\newline
    \ - Describe the goal of the task and provide concrete information or constraints. Use mock-up information (identifier, number, personal information, name, date, attributes, constraints, etc.) to make the task more specific and realistic.}\\\midrule
        \textbf{Sample Responses}&Retrieve the confirmation number for a reservation made under the name jane smith departing from dallas (dfw) to miami (mia) on january 20th.\\\addlinespace
            &Find and book a round-trip flight from new york to london departing on december 15th and returning on december 22nd for under \$800.\\\addlinespace
            &Find the lowest fare for a one-way flight from new york city (jfk) to los angeles (lax) leaving on august 15th and returning on august 20th, with no layovers exceeding 2 hours.\\\addlinespace
            &Rent a car in honolulu for one week starting on august 1st that can fit at least four passengers and has gps navigation included.\\\addlinespace
            &Cancel a car rental reservation for a compact car in orlando, florida that was scheduled to be picked up on may 25th at 5 pm without incurring any cancellation fees.\\
         \bottomrule
    \end{tabular}
    \label{tab:task_generation_prompt}
\end{table}

In this study, we employ annotators from the Amazon Mechanical Turk (mturk) platform. The annotators are required to have a minimum of \num{1000} approved HITs with an approval rate exceeding \num{98}\% on the platform.
We design the compensation with an estimated hourly rate of \$10.10 to respect the minimum wage guidelines in Ohio, United States.
Every worker passing our qualification receives a bonus, and we pay \$\num{0.80} for each approved final task.
We do not collect any identifiable private information during the study, and explicitly instruct the annotators to refrain from entering personal or sensitive data into the system. Annotators engage with our annotation tool only within a secure, remote sandbox environment, posing no foreseeable harm.
The study complies with the IRB exemption criteria, per the Office of Responsible Research Practices at The Ohio State University. All annotators are presented with a consent form, to which they must agree before participating in the study.
To prepare the workers for the task, we provide a comprehensive training document and a video tutorial, followed by a qualification assessment comprising a questionnaire and a series of test demonstrations using our tool.
It is noteworthy that the task is divided into two phases: task proposal and task demonstration. The proposal phase comes with a nominal reward, with the majority of the compensation dispensed upon successful completion of the demonstration.

Quality and diversity are ensured through a two-stage review process.
The first author reviews all tasks after task proposal and manually select the tasks for demonstration.
After task demonstration, a thorough final verification of all collected data is conducted by all authors to authenticate the tasks and recorded actions.
Each demonstration is first verified by one of the authors, and uncertain ones are further verified by the first author to reach consensus.

\begin{figure}
    \centering
    \includegraphics[width=\linewidth]{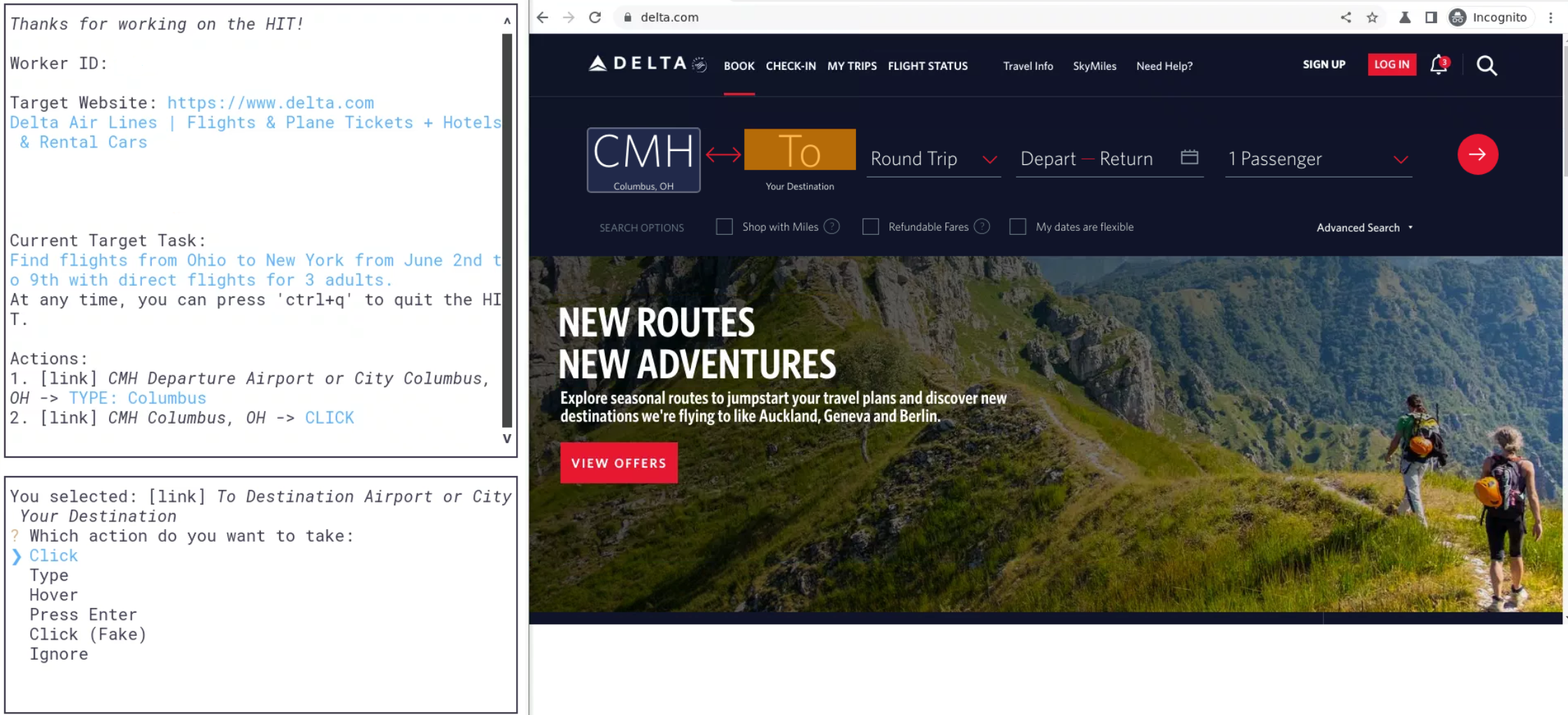}
    \caption{Illustration of our annotation tool, which consists of two side-by-side windows. On the left we provide a dialogue window for the user to control the tool and select operations to take. On the right we provide the browser window for the user to interact with and select web elements.}
    \label{fig:tool_0}
\end{figure}

\subsection{Task Proposal}
We use ChatGPT to generate sample tasks to provide inspiration to the annotators, and the prompt used is shown in Table~\ref{tab:task_generation_prompt}.
For each HIT, we ask the annotator to select a website of their interest first.
Following this, we present them with ten sample tasks produced by ChatGPT, and request them to propose a maximum of five additional tasks.
The annotator is instructed not to directly copy the sample tasks.
We manually evaluate all submitted tasks and reject those that demonstrate low quality or are too similar to previously accepted tasks.
We set a nominal reward of \$\num{0.05} for each task proposal HIT, and the annotator will receive it no matter whether the tasks are accepted or not. For accepted ones, the annotator will receive the full reward of \$\num{0.80} on successful demonstration of the task.
Once we have collected a total of around \num{20} tasks for a specific website, we desist from showing it to the user, aiming for a balanced distribution among websites and increased task diversity.

\subsection{Task Demonstration}
\begin{figure}
    \centering
    \includegraphics[width=.85\linewidth]{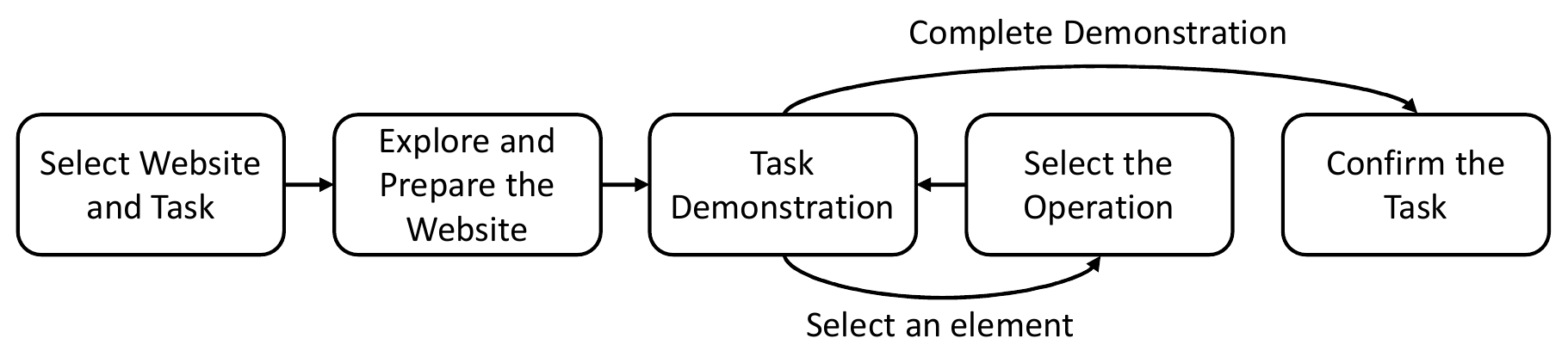}
    \caption{The overall procedure for task demonstration.}
    \label{fig:tool_flow}
\end{figure}
\begin{figure}
\begin{minipage}[t]{.56\linewidth}
       \centering
        \includegraphics[width=0.9\textwidth]{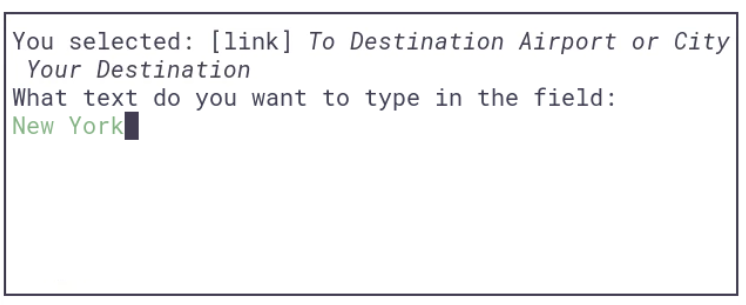}
        \caption{Dialogue window for the \texttt{Type} operation.}
        \label{fig:tool_type}
    \end{minipage}%
    \hfill
    \begin{minipage}[t]{.42\linewidth}
        \centering
        \includegraphics[width=0.93\textwidth]{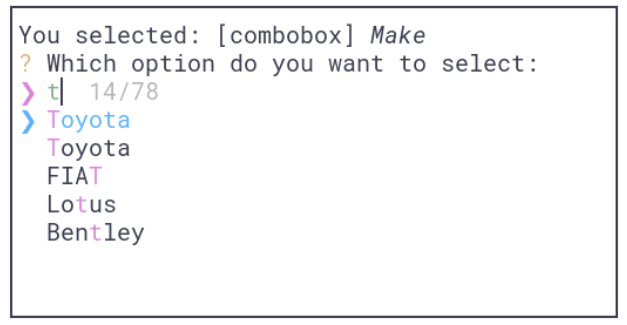}
        \caption{Dialogue window for the \texttt{Select Option} operation.}
        \label{fig:tool_select}
    \end{minipage} 
\end{figure}

We develop a dedicated annotation tool for task demonstration using Playwright,\footnote{\url{https://playwright.dev/}} which allows us to interact with the browser and record user actions.
As shown in Figure~\ref{fig:tool_0}. the tool is composed of two windows.
The dialogue window on the left serves as the annotator's control panel for guiding the interaction flow and choosing operations.
The browser window on the right is where the annotator navigates the website and selects elements for interaction.
Figure~\ref{fig:tool_flow} shows the overall procedure for task demonstration.
The annotator starts by selecting the website and task to be demonstrated. Once selected, the tool will bring up the website in the browser.
The annotator is then instructed to explore the website and practice the task.
To collect clean actions during actual demonstration, the workers are asked to close pop-up windows during exploration. We also provide anonymous accounts for the workers to use so that no private information is entered.
The exploration stage is not recorded and primarily serves to familiarize the annotator with the website and task, as well as to prepare the website to prevent future pop-ups, thereby ensuring a clean, streamlined set of final recorded actions.
After exploration, the annotator is directed to return to the homepage and reset any altered values, allowing us to begin the demonstration in a fresh state.
During the demonstration, the annotator will illustrate how to accomplish the task step-by-step using both the browser and the dialogue window. 
To ensure a clean set of annotated actions, annotators are restricted from directly engaging with the browser during the demonstration phase.
Instead, we divide each action step into two stages: Element selection and operation selection.
At each step, the annotator first selects the target element by clicking it in the browser. We will highlight the selected element in the browser window but block the actual click event.
The annotator is then prompted to select the operation to perform within the dialogue window, which is then carried out by the annotation tool in the browser.
We provide \num{6} operations: \texttt{Click}, \texttt{Type}, \texttt{Hover}, \texttt{Press Enter}, \texttt{Click (Fake)} and \texttt{Ignore}.
For the \texttt{Type} operation, the annotator is additionally required to supply the value as shown in Figure~\ref{fig:tool_type}.
If the chosen element is a \texttt{select} HTML element, and the annotator opts for \texttt{Click}, it translates to a \texttt{Select Option} operation and we will prompt the annotator to select one of the options as shown in Figure~\ref{fig:tool_select}.
To avoid ambiguity, the \texttt{Click}, \texttt{Hover} and \texttt{Press Enter} operations are all mapped to \texttt{Click} in the final dataset.
\texttt{Click (Fake)} is a special operation. It will be recorded the same as a normal \texttt{Click} but will not get executed in the browser. This is designed for safeguarding against state-changing actions (i.e., actions that produce side effects to the world), such as posting a comment or scheduling an appointment, since it will interfere with other real users of the website.
In practice, once a model predicts \texttt{Click (Fake)}, it may prompt the user for confirmation before executing such state-changing actions.
Finally, the annotator can also choose \texttt{Ignore} in case they select a wrong element.
Once all the actions have been annotated, the annotator can choose to complete the task. They will then be asked to confirm the task description again and make any necessary modifications.

\noindent\textbf{Pop-ups and CAPTCHAs.}
In this study, we emphasize on clean and direct task execution actions, intentionally omitting extraneous steps like pop-ups and CAPTCHAs that might introduce ambiguity in evaluation. We carefully select only those websites that pose no access issues when used with our tool.
Before recording the task demonstration, annotators are requested to familiarize themselves with the website, and preemptively close pop-up windows and clear CAPTCHAs to avoid their recurrence during the actual demonstration.
Annotators are further guided not to engage in extra steps such as closing ads unless necessary during the task demonstration.
In the final task verification, we revisit the actions and filter out those unrelated to direct task execution.
At the same time, we acknowledge that these instances constitute a significant aspect of the dynamic web environment in the real world.
Enhancing systems to robustly tackle such scenarios on-the-go could form an interesting avenue for future research.

\noindent\textbf{Mitigating Disruptions on the Websites.}
The annotator are advised against actions that could potentially interfere the normal operation of the website. 
To handle tasks such as scheduling appointments, we introduce a \texttt{Click (Fake)} operation that annotators can utilize to indicate the action without actually executing it on the website.

\subsection{Task Verification}
\begin{figure}
    \centering
    \includegraphics[width=\linewidth]{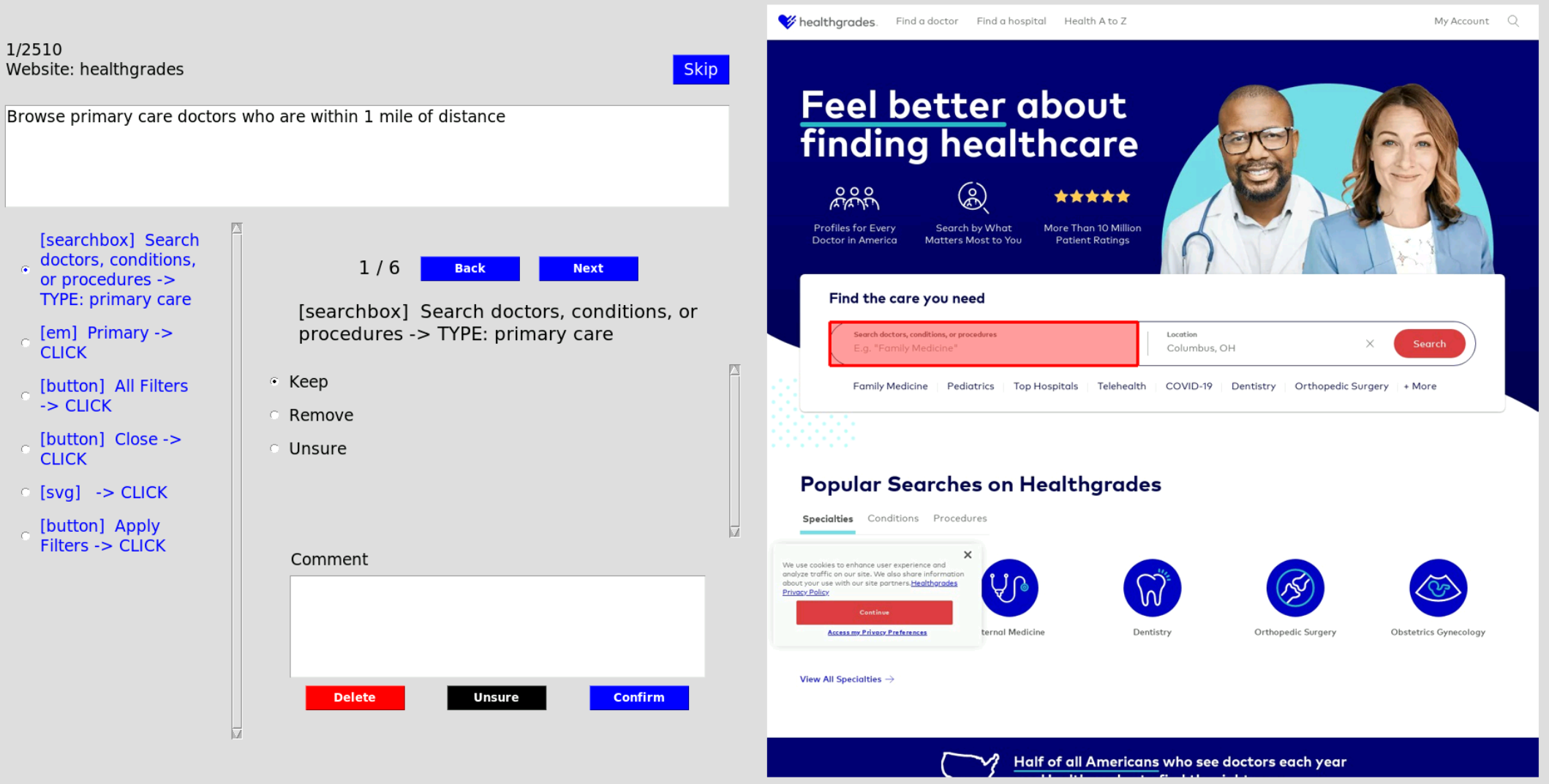}
    \caption{Illustration of our verification tool.}
    \label{fig:tool_verification}
    \vspace{-10pt}
\end{figure}
All collected data undergoes an additional verification process conducted by the authors, as demonstrated in Figure~\ref{fig:tool_verification}. The verification interface is shown in Figure~\ref{fig:tool_verification}. This verification consists of three tasks.
Firstly, we evaluate whether a task should be discarded due to its low quality. 
Secondly, we examine each step to determine if any action should be discarded. This includes reviewing the initial and final actions of the task, and excluding any additional actions (e.g., closing ads) that are not outlined in the task description to ensure consistency across task annotations.
Finally, we verify the task description to confirm that all actions are accurately represented and make modifications if necessary.
If there is uncertainty regarding any action, the verifier can opt for the `unsure' option, prompting a re-evaluation by the first author.
\section{Experiment Details}
\label{app:exp}
\subsection{Evaluation}
\label{app:eval_normalization}
One complication that arises during evaluation on real-world websites is that multiple elements on a webpage may induce the same effect. For instance, a button might house a text span within it, both of which, when clicked, yield identical results.
To enhance the robustness of our evaluation, we employ heuristics to detect elements equivalent to the ground truth.
We first examine the ancestors of the labeled element to identify potential higher-level elements acceptable for the current action. We employ a straightforward heuristic that locates the nearest clickable element
to the ground truth, including itself. After identifying the top-level acceptable element, we include all its visible descendants that are located within its post-rendering bounding box as acceptable as well. Manual checking on \num{100} instances where the heuristic identifies a top-level element other than the ground truth confirms the validity of the approach.
 For both training and evaluation stages, all acceptable elements are considered positive.

\subsection{Model Implementation Details}

\begin{table}[t]
    \centering
    \small
    \caption{Hyperparameters used in experiments.}
    \begin{tabular}{rp{3cm}p{6cm}}
    \toprule
        \textbf{Method} & \textbf{Model} & \textbf{Hyperparamerters} \\
        \midrule
        Candidate Generation & \texttt{deberta-v3-base} &\texttt{batch\_size:\num{32}, epoch:\num{5}, learning\_rate:\num{3e-5}}\\\addlinespace
        Action Prediction \\
        \hspace{2em}Generation& \texttt{flan-t5-base} &\texttt{batch\_size:\num{32}, epoch:\num{5}, learning\_rate:\num{5e-5}}\\\addlinespace
        \multirow{1}{*}{\hspace{2em}\model}& \texttt{flan-t5-base, flan-t5-large, flan-t5-xl} &\texttt{batch\_size:\num{32}, epoch:\num{5}, learning\_rate:\num{5e-5}}\\
        &\texttt{gpt-3.5-turbo, gpt-4}&\texttt{temperature:\num{0}, \#~demonstrations:\num{3}}\\
    \bottomrule
    \end{tabular}
    \label{tab:hyper}
\end{table}

\noindent\textbf{Candidate Generation.}
We use the Cross-Encoder implementation from Sentence-Transformers~\footnote{\url{https://www.sbert.net/examples/applications/cross-encoder/README.html}} and use DeBERTa as the backbone model. More specifically, we use \texttt{DeBERTa-v3-base}~\footnote{\url{https://huggingface.co/microsoft/deberta-v3-base}} for our experiments.

\noindent\textbf{Action Prediction.}
We use the Seq2Seq model implementation from Transformers~\cite{wolf-etal-2020-transformers}. We experiment with the \texttt{base}~\footnote{\url{https://huggingface.co/google/flan-t5-base}}, \texttt{large}~\footnote{\url{https://huggingface.co/google/flan-t5-large}} and \texttt{xl}~\footnote{\url{https://huggingface.co/google/flan-t5-xl}} versions of Flan-T5~\cite{flant5}.

\noindent\textbf{LLM In-context Learning.}
We use the OpenAI API for in-context learning with LLMs. We experiment with two versions of GPT models: \texttt{gpt-3.5-turbo} and \texttt{gpt-4}. We include three demonstration examples for in-context learning. The complete prompt is shown in Table~\ref{tab:action_prediction_prompt}.

\noindent\textbf{Training Details.}
The \texttt{flan-t5-xl} and \texttt{flan-t5-large} models are trained on servers with 4*A100 80GB cards provided by Ohio Supercomputer Center~\cite{OhioSupercomputerCenter1987}. All other models are trained with single A6000 48GB cards.

Please see Table~\ref{tab:hyper} for all hyperparameters used in our experiments.


\section{Additional Results}
\label{app:add_result}
\subsection{Effect of Random Grouping Elements for Action Prediction}
\label{app:random_group}
\begin{table}[]
    \centering
    \caption{Step Success Rate for Flan-T5 models with different groups of options. Here we shown mean and standard deviation of 5 runs with different random seeds.}
    \begin{tabular}{lccc}
    \toprule
         & \textbf{Cross-Task}& \textbf{Cross-Website} & \textbf{Cross-Domain} \\
         \midrule
      Flan-T5\textsubscript{B}   &  \num{41.5}$\pm$\num{0.7}&\num{30.0}$\pm$\num{0.8}&\num{31.3}$\pm$\num{0.5}\\
      Flan-T5\textsubscript{L}   &  \num{49.9}$\pm$\num{0.2}&\num{35.7}$\pm$\num{0.5}&\num{36.7}$\pm$\num{0.3}\\
      Flan-T5\textsubscript{XL}   &  \num{51.9}$\pm$\num{0.8}&\num{39.5}$\pm$\num{0.2}&\num{39.6}$\pm$\num{0.2}\\
      \bottomrule
    \end{tabular}
    \label{tab:random_runs}
\end{table}
For both training and inference, we shuffle the elements in the webpage and randomly group them into multi-choice questions. The model might give different predictions when presented with different sets of choices, and leads to slightly different final evaluation scores. Here we show the average and standard deviation of \num{5} runs with different random seeds to show the effect of random grouping. As we can see from Table~\ref{tab:random_runs}, the selection of choices only lead to small changes in overall performance with standard deviation less than 1 for all runs.
\subsection{Zero-shot Results for Flan-T5\textsubscript{XL}}
\label{app:flan-t5_zero}
\begin{table}[]
    \centering
    \caption{Zero-shot element selection results for Flan-T5\textsubscript{XL} compared with the fine-tuned counterpart.}
    \begin{tabular}{lccc}
         \toprule
         & \textbf{Cross-Task}& \textbf{Cross-Website} & \textbf{Cross-Domain} \\
         \midrule
         Flan-T5\textsubscript{XL} Zero-Shot& \num{10.8} &	\num{7.8} & \num{11.7} \\
         Flan-T5\textsubscript{XL} Fine-Tuned& \num{52.0} &\num{38.9}&\num{39.6}\\
         \bottomrule
    \end{tabular}
    \label{tab:flan-t5_zero}
\end{table}
Since Flan-T5 is tuned with multi-choice format, it can also do element selection in zero-shot. However, as we can see from Table~\ref{tab:flan-t5_zero}, while the model still gets some elements correct, it is much lower compared to the fine-tuned model, and \num{3}-shot GPT 3.5/4. This is expected, since Flan-T5 is not tuned for HTML and coding related tasks.
\subsection{Results on the \num{50} task subsets}
\label{app:result_50}
\begin{table}[]
    \centering
        \caption{Step Success Rate for all methods on the \num{50} tasks subsets we used to evaluate GPT-4. Numbers in parentheses are the results on the full test set (same as Table~\ref{tab:main})}
    \begin{tabular}{lccc}
         \toprule
         & \textbf{Cross-Task}& \textbf{Cross-Website} & \textbf{Cross-Domain} \\
         \midrule
Flan-T5\textsubscript{B}&	\num{43.3} (\num{41.0})&	\num{25.3} (\num{29.5})&	\num{28.1} (\num{31.6})\\
Flan-T5\textsubscript{L}&	\num{48.1} (\num{50.3})&	\num{30.8} (\num{35.3})&	\num{27.6} (\num{37.3})\\
Flan-T5\textsubscript{XL}&	\num{47.9} (\num{52.0})&	\num{33.3} (\num{38.9})&	\num{34.6} (\num{39.6})\\
GPT-3.5&	\num{15.2} (\num{17.4})&	\num{15.1} (\num{16.2})&	\num{16.7} (\num{18.6})\\
GPT-4&	\num{36.2}&	\num{30.1}&	\num{26.4}\\
\bottomrule
    \end{tabular}
    \label{tab:result_50}
\end{table}

Due to budget constraint, we only run GPT-4 on \num{50} tasks for each setting. Here we show the step success rate results for other methods on the same \num{50} examples that GPT-4 is tested on,  
As we can see from Table~\ref{tab:result_50}, the results on the \num{50} tasks subsets are consistent with the results on the respective full test set, and the relative performance across methods and splits remains the same.

\newpage
\begin{center}
    \begin{longtable}{lp{11.5cm}}
    \caption{Prompt for action prediction in \model\ with GPT models. Only part of the HTML snippet is shown here to save space. 
    }
    \label{tab:action_prediction_prompt}\\

\toprule \multicolumn{1}{l}{\textbf{Role}} & \multicolumn{1}{c}{\textbf{Content}} \\ \midrule
\endfirsthead

\multicolumn{2}{c}%
{{\bfseries \tablename\ \thetable{} -- continued from previous page}} \\
\midrule \multicolumn{1}{l}{\textbf{Role}} & \multicolumn{1}{c}{\textbf{Content}} \\ \midrule 
\endhead

\midrule \multicolumn{2}{r}{{Continued on next page}} \\ \bottomrule
\endfoot

\endlastfoot
    system & You are a helpful assistant that is great at website design, navigation, and executing tasks for the user\\\addlinespace
    user&
    \textasciigrave\textasciigrave\textasciigrave\newline
    <html> <div> <div> <a tock home page /> <button id=0 book a reservation. toggle open> <span> Book a reservation </span> </button> <button book a reservation. toggle open> </button> </div> <div> <select id=1 type> <option reservations true> Dine in </option> ... </html>\newline
    \textasciigrave\textasciigrave\textasciigrave\newline
    Based on the HTML webpage above, try to complete the following task:\newline
    Task: Check for pickup restaurant available in Boston, NY on March 18, 5pm with just one guest\newline
    Previous actions:\newline
    None\newline
    What should be the next action? Please select from the following choices (If the correct action is not in the page above, please select A. `None of the above'):\newline\newline
    A. None of the above\newline
    B. <button id=0 book a reservation. toggle open> <span> Book a\newline
    C. <select id=1 type> <option reservations true> Dine in </option> <option\newline
    D. <div id=2> <p> Celebrating and supporting leading women shaking up\\\addlinespace
    assistant&
    Answer: C.\newline
    Action: SELECT\newline
    Value: Pickup\\\addlinespace
    user&
    \textasciigrave\textasciigrave\textasciigrave\newline
    <html> <div> <main main> <section tabpanel> <div> <ul tablist> <li tab heading level 3 search and> </li> <li id=0 tab heading level 3 search and> <span> Hotel </span> </li> <li tab heading level 3 search and> </li> <li tab heading level 3 search and> </li> </ul> <div tabpanel> <div id=1> <div> <span> Dates* </span> <button button clear dates /> </div> ... </html>\newline
    \textasciigrave\textasciigrave\textasciigrave\newline
    Based on the HTML webpage above, try to complete the following task:\newline
    Task: Compare the fare types to book a 1-adult ticket from Springfiels, IL to Austin, TX for April 29th 2023\newline
    Previous actions:\newline
    [combobox]  Enter your departing city, airport name, or airpor... -> TYPE: SPRINGFIELD\newline
    [button]  Springfield, IL, US (SPI) -> CLICK\newline
    [combobox]  Enter your destination city, airport name, or airp... -> TYPE: AUSTIN\newline
    [button]  Austin, TX, US (AUS) -> CLICK\newline
    What should be the next action? Please select from the following choices (If the correct action is not in the page above, please select A. `None of the above'):\newline\newline
    A. None of the above\newline
    B. <li id=0 tab heading level 3 search and> <span> Hotel\newline
    C. <div id=1> <div> <span> Dates* </span> <button button clear dates\newline
    D. <ul id=2> <a mobile tools> </a> <a open united's tiktok\\\addlinespace
    assistant&Answer: A.\\\addlinespace
    user&
    \textasciigrave\textasciigrave\textasciigrave\newline
    <html> <div> <nav main menu> <ul> <li> <div button> Car Sales </div> <div id=0> <div> <div> <div> Buy A Car </div> <div> Plan Your Purchase </div> </div> <div> <h4> Its Tax Refund Time. Treat Yourself to an Upgrade. </h4> <p> With a variety of options, invest your refund in what you really want - a quality, used vehicle from Enterprise. </p> ... </html>\newline
    \textasciigrave\textasciigrave\textasciigrave\newline\newline
    Based on the HTML webpage above, try to complete the following task:\newline
    Task: Find a mini van at Brooklyn City from April 5th to April 8th for a 22 year old renter.\newline
    Previous actions:\newline
    [searchbox]  Pick-up \& Return Location (ZIP, City or Airport) (... -> TYPE: Brooklyn\newline
    [option]  Brooklyn, NY, US Select -> CLICK\newline
    What should be the next action? Please select from the following choices (If the correct action is not in the page above, please select A. `None of the above'):\newline\newline
    A. None of the above\newline
    B. <div id=0> <div> <div> <div> Buy A Car </div> <div>\newline
    C. <div id=1> Enterprise Fleet Management </div>\newline
    D. <button id=2 selected pick-up date 03/19/2023> <span> <span> 19 </span>\\\addlinespace
    assistant&
    Answer: D.\newline
    Action: CLICK\\
    \bottomrule
    \end{longtable}
\end{center}

\end{document}